\pdfoutput=1

\documentclass[11pt]{article}

\usepackage{emnlp2021}
\usepackage{times}
\usepackage{latexsym}
\usepackage{color}
\usepackage{graphicx}
\usepackage{amsmath}
\usepackage{amsfonts}
\usepackage[T1]{fontenc}
\usepackage{amsfonts}
\usepackage{multirow}
\usepackage{color}
\usepackage[utf8]{inputenc}

\usepackage{microtype}
\newcommand{\mytext}[1]{{\textbf{\textit{#1}}}}
\title{Importance Estimation from Multiple Perspectives for \\ Keyphrase Extraction}

\author{Mingyang Song,  Liping Jing\thanks{\; Corresponding author.} \ and Lin Xiao\\
         Beijing Key Lab of Traffic Data Analysis and Mining \\
         Beijing Jiaotong University, China \\
         \texttt{\{mingyang.song, lpjing, 17112079\}@bjtu.edu.cn}}

\begin{document}
\maketitle
\begin{abstract}
Keyphrase extraction is a fundamental task in Natural Language Processing, which usually contains two main parts: candidate keyphrase extraction and keyphrase importance estimation. From the view of human understanding documents, we typically measure the importance of phrase according to its syntactic accuracy, information saliency, and concept consistency simultaneously. However, most existing keyphrase extraction approaches only focus on the part of them, which leads to biased results. In this paper, we propose a new approach to estimate the importance of keyphrase from multiple perspectives (called as \textit{KIEMP}) and further improve the performance of keyphrase extraction. Specifically, \textit{KIEMP} estimates the importance of phrase with three modules: a chunking module to measure its syntactic accuracy, a ranking module to check its information saliency, and a matching module to judge the concept (i.e., topic) consistency between phrase and the whole document. These three modules are seamlessly jointed together via an end-to-end multi-task learning model, which is helpful for three parts to enhance each other and balance the effects of three perspectives. Experimental results on six benchmark datasets show that \textit{KIEMP} outperforms the existing state-of-the-art keyphrase extraction approaches in most cases.
\end{abstract}

\section{Introduction}
Keyphrase Extraction (KE) aims to select a set of reliable phrases (e.g., ``harmonic balance method", ``grobner base", ``error bound", ``algebraic representation", and ``singular point" in Table~\ref{kg}) with salient information and central topics from a given document, which is a fundamental task in natural language processing.
Most classic keyphrase extraction methods typically include two mainly components: \textit{candidate keyphrase extraction} and \textit{keyphrase importance estimation} \cite{medelyan2009,2010topic,2014survey}. 

\begin{table}[!t]
	\renewcommand\arraystretch{1.5}
	\begin{center}
		\small
		\begin{tabular}{|p{7.2cm}|}
			\hline
			\textit{Input Document}:
			
			
			\mytext{\textcolor{magenta}{harmonic balance ( hb ) method}} is well known principle for analyzing periodic oscillations on nonlinear networks and systems. because the hb method has a truncation error, approximated solutions have been guaranteed by \mytext{\textcolor{magenta}{error bounds}}. however, its numerical computation is very time consuming compared with solving the hb equation. this paper proposes proposes an \mytext{\textcolor{magenta}{algebraic representation}} of the \mytext{\textcolor{magenta}{error bound}} using \mytext{\textcolor{magenta}{grobner base}}. the \mytext{\textcolor{magenta}{algebraic representation}} enables to decrease the computational cost of the \mytext{\textcolor{magenta}{error bound}} considerably. moreover, using \mytext{\textcolor{magenta}{singular points}} of the \mytext{\textcolor{magenta}{algebraic representation}}, we can obtain accurate break points of the \mytext{\textcolor{magenta}{error bound}} by collisions.
			\\ \hline
			\textit{Output / Target Keyphrases}: 
			
			\mytext{\textcolor{magenta}{harmonic balance method}}; \mytext{\textcolor{magenta}{grobner base}}; \mytext{\textcolor{magenta}{error bound}}; \mytext{\textcolor{magenta}{algebraic representation}}; \mytext{\textcolor{magenta}{singular point}}; \mytext{\textcolor{cyan}{quadratic approximation}}
			\\\hline
		\end{tabular}
		\caption{\label{kg} Sample input document with output / target keyphrases in \textit{KP20k} testing set. Specially, keyphrases typically can be categorized into two types: \mytext{\textcolor{magenta}{present keyphrase}} that appears in a given document and \mytext{\textcolor{cyan}{absent keyphrase}} which does not appear in a given document.}
	\end{center}
\end{table}

As shown in Table~\ref{kg}, each keyphrase usually consists of more than one words \cite{catseq17}. To extract the candidate keyphrases from the the given document which is typically characterized via word-level representation, researchers leverage some heuristics \cite{heuristic1,heuristic_liu_a,heuristic_liu_b,heuristic3,heuristic5,medelyan2009} to identify the candidate keyphrases.
For example, the word embeddings are composed to n-grams by Convolution Neural Network (CNN) \cite{xiong19,baseline, 2020sota}.

Usually, the candidate set contains much more keyphrases than the ground truth keyphrase set.
Therefore, it is critical to select the important keyphrase from the candidate set by a good strategy. 
In other words, \textit{keyphrase importance estimation} commonly is one of the essential components in many keyphrase extraction models.
Since keyphrase extraction concerns “the automatic selection of important and topical phrases from the body of a document” \cite{turney2000}. Its goal is to estimate the importance of the candidate keyphrases to determine which one should be extracted.
Recent approaches \cite{baseline, 2020sota} recast the keyphrase extraction as a classification problem, which extracts keyphrases by a binary classifier.
However, a binary classifier classifies each candidate keyphrase independently, and consequently, it does not allow us to determine which candidates are better than the others \cite{hulth2004}. 
Therefore, some methods \cite{jiang2009,xiong19,2020sota,baseline} propose a ranking model to extract keyphrases, where the goal is to learn a phrase ranker to compare the saliency of two candidate phrases.
Furthermore, many previous studies \cite{2010topic, topic, heuristic_liu_b} extract keyphrases with the main topics discussed in the source document, 
For example, \citet{2010topic} proposes to build a topical PageRank approach to measure the importance of words concerning different topics.

However, most existing keyphrase extraction methods estimate the importance of keyphrases on at most two perspectives, leading to biased extraction.
Therefore, to improve the performance of keyphrase extraction, the importance of the candidate keyphrases requires to be estimated sufficiently from multiple perspectives.
Motivated by the phenomenon mentioned above, we propose a new importance estimation from multiple perspectives simultaneously for the keyphrase extraction task.
Concretely, it estimates the importance from three perspectives with three modules (syntactic accuracy, information saliency, and concept consistency) with three modules.
A chunking module, as a binary classification layer, measures the syntactic accuracy of each candidate keyphrase.
A ranking module checks the semantics saliency of each candidate phrase by a pairwise ranking approach, which introduces competition between the candidate keyphrases to extract more salient keyphrases.
A matching module judges the concept relevance of each candidate phrase in the document via a metric learning framework.
Furthermore, our model is trained jointly on the above three modules, balancing the effect of three perspectives.
Experimental results on two benchmark data sets show that \textit{KIEMP} outperforms the existing state-of-the-art keyphrase extraction approaches in most cases.

\section{Related Work}
A good keyphrase extraction system typically consists of two steps: (1) \textit{candidate keyphrase extraction}, extracting a list of words / phrases that serve as the candidate keyphrases using some heuristics \cite{heuristic1,heuristic3,medelyan2009,heuristic5,heuristic_liu_a,heuristic_liu_b}; and (2) \textit{keyphrase importance estimation}, determining which of these candidate phrases are keyphrases using different importance estimation approaches.

In the \textit{candidate keyphrase extraction}, the heuristic rules usually are designed to avoid spurious phrases and keep the number of candidates to a minimum \cite{2014survey}.
Generally, the heuristics mainly include (1) leverage a stop word list \cite{heuristic_liu_b}, (2) allowing words with part-of-speech tags \cite{textrank,heuristic_liu_a}, (3) composing words to n-grams to be the candidate keyphrases \cite{medelyan2009, baseline, xiong19, 2020sota}.
The above heuristics have proven effective with their high recall in extracting gold keyphrases from various sources.
Motivated by the above methods, in this paper, we leverage CNNs to compose words to n-grams as the candidate keyphrases.

In the \textit{keyphrase importance estimation}, the existing methods can be mainly divided into two categories: unsupervised and supervised.
The unsupervised method usually are categorized into four groups, i.e., graph-based ranking \cite{textrank}, topic-based clustering \cite{heuristic_liu_b}, simultaneous learning \cite{zha2002}, and language modeling \cite{tomokiyo2003}.
Early supervised approaches to keyphrase extraction recast this task as a binary classification problem \cite{frank1999,turney1999,turney2000,jiang2009}.
Later, to determine which candidates are better than the others, many ranking approach is proposed to rank the saliency of two phrases \cite{jiang2009,baseline}. This pairwise ranking approach, therefore, introduces competition between candidate keyphrases and has been achieved good performance.
Both supervised and unsupervised methods construct features or models from different perspectives to measure the importance of candidate keyphrases to determine which keyphrases should be extracted.
However, the approaches mentioned earlier consider at most two perspectives when measuring the importance of phrases, which leads to biased keyphrase extraction.
Different from the existing methods, the proposed \textit{KIEMP} considers estimating the importance of the candidate keyphrases from multiple perspectives simultaneously.

\section{Methodology}
We formally define the problem of keyphrase extraction as follows.
In this paper, \textit{KIEMP} takes a document ${D}=\{w_1, ..., w_i, ..., w_M\}$ and learns to extract a set of keyphrases ${K}$ (each keyphrase may be composed of one or several word(s)) from their n-gram based representations under multiple perspectives.

\begin{figure*}
	\centering
	\includegraphics[scale=0.5]{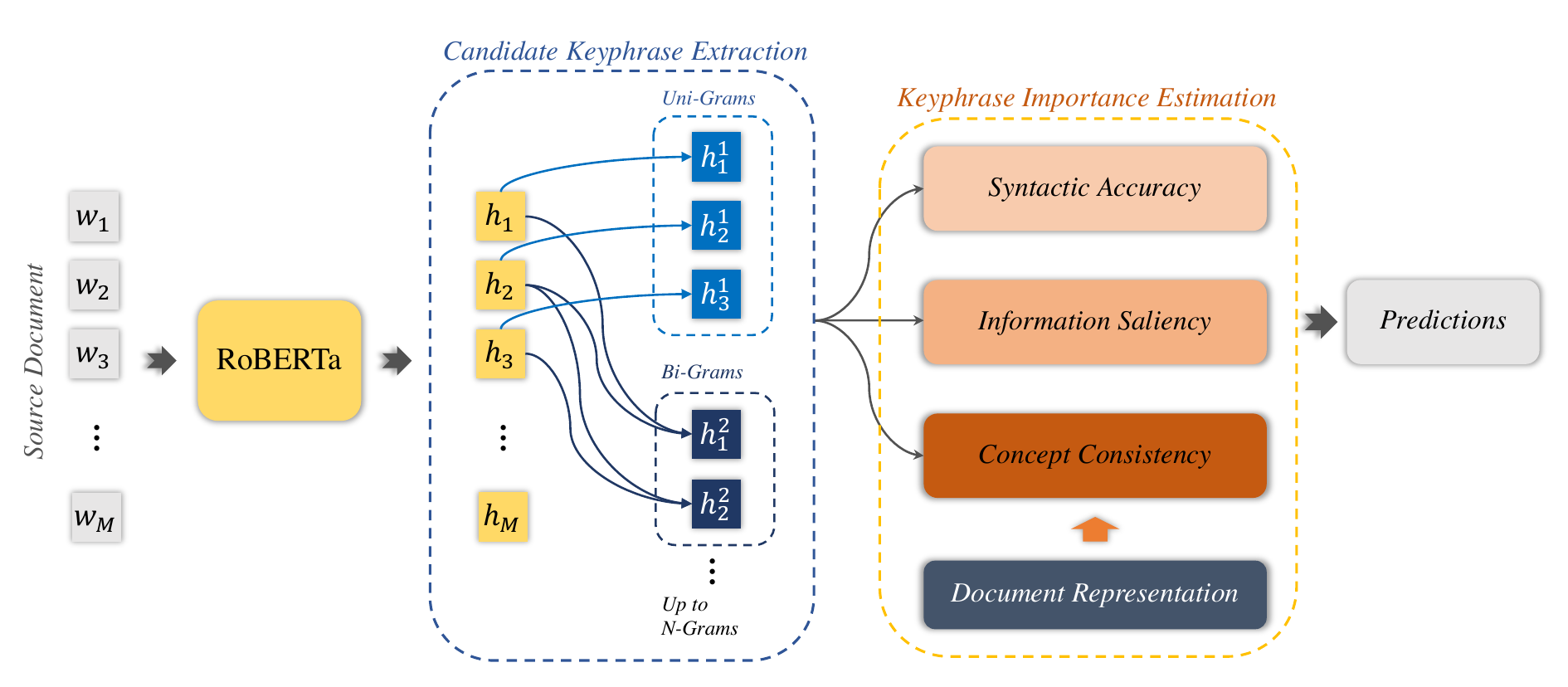}
	\caption{The \textit{KIEMP} model architecture. }
	\label{fig:model}
\end{figure*}

This section describes the architecture of \textit{KIEMP}, as shown in Figure~\ref{fig:model}.
\textit{KIEMP} mainly consists of two submodels: candidate keyphrase extraction and keyphrase importance estimation.
The former first identifies and extracts the candidate keyphrases. Then the latter estimates the importance of keyphrases from three perspectives simultaneously with three modules to determine which one should be extracted.

\subsection{Contextualized Word Representation}
Recently, pre-trained language models \cite{elmo, bert,roberta} have emerged as a critical technology for achieving impressive gains in a wide variety of natural language tasks \cite{bert_summary}.
These models extend the idea of word embeddings by learning contextual representations from large-scale corpora using a language modeling objective.
In this situation, \citet{xiong19} propose to represent each word by its ELMo \cite{elmo} embedding and \citet{baseline} leverage variants of BERT \cite{bert, roberta} to obtain contextualized word representations.
Motivated by the above approaches, we represent each word by RoBERTa \cite{roberta}, which encodes ${D}$ to a sequence of vector ${H}=\{h_1, ..., h_i, ..., h_M\}$:
\begin{equation}
	{H}= \text{RoBERTa} \{w_1, ..., w_i, ..., w_M\},
\end{equation}
where $h_i \in \mathbb{R}^d$ indicates the $i$-th contextualized word embedding of $w_i$ from the last transformer layer in RoBERTa.
Specifically, the [CLS] token of RoBERTa is used as the document representation.
\subsection{Candidate Keyphrase Extraction}
In the keyphrase extraction task, keyphrase usually contains more than one word, as shown in Table~\ref{kg}.
Therefore, it is necessary to identify the candidate keyphrases via some strategies.
Previous work \cite{medelyan2009,baseline,2020sota,xiong19} allow n-grams that appear in the document to be the candidate keyphrases.
Motivated by the previous approaches, we consider the language properties \cite{xiong19} and compose the contextualized word representations to n-grams by CNNs (similar to \citet{baseline}).
Specifically, the phrase representation of the $i$-th $n$-gram $c_i^n$ is computed as:
\begin{equation}
	h_i^n = \text{CNN}^n (h_{i:i+n}),
\end{equation}
where $h_i^n \in \mathbb{R}^d$ indicates the $i$-th $n$-gram representation.
Concretely, $n\in[1, N]$ is the length of n-grams, and $N$ indicates the maximum length of allowed candidate n-grams.
Specifically, each n-gram has its own set of convolution filters $\text{CNN}^n$ with window size $n$ and stride $1$.

\subsection{Keyphrase Importance Estimation}
In the keyphrase extraction models, \textit{keyphrase importance estimation} commonly is one of the essential components.
To improve the accuracy of keyphrase extraction, we estimate the importance of keyphrases from three perspectives simultaneously with three modules: chunking for syntactic accuracy, ranking for information saliency, and matching for concept consistency.
\subsubsection{Chunking for Syntactic Accuracy} 
Many studies \cite{turney1999,frank1999,turney2000} regard keyphrase extraction as a classification task, in which a model is trained to determine whether a candidate phrase is a keyphrase in a syntactic perspective.
For example, \citet{xiong19, baseline} directly predict whether the n-gram is a keyphrase based on its corresponding representation.
Motivated by these above methods, in this paper, the syntactic accuracy of phrase $c_i^n$ is estimated by a chunking module:
\begin{equation}
	I_1(c_i^n) = \text{softmax}(\mathbf{W}_1h_i^n + b_1),
\end{equation}
where $\mathbf{W}_1$ and $b_1$ indicate a trainable matrix and a bias.
The softmax is taken over all possible n-grams at each position $i$ and each length $n$.
The whole model is trained using cross-entropy loss:
\begin{equation}
	L_c = \text{CrossEntropy}(y_i^n, I_1(c_i^n)),
\end{equation}
where $y_i^n$ is the label of whether the phrase $c_i^n$ is a keyphrase of the original document.
\subsubsection{Ranking for Information Saliency}
The binary classifier-based keyphrase extraction model classifies each candidate keyphrase independently, and consequently, it does not allow us to determine which candidates are better than the others \cite{hulth2004}. 
However, the goal of keyphrase extraction is to identify the most salient phrases for a document \cite{2014survey}.
Therefore, a ranking model is required to rank the saliency of the candidate keyphrases.
We leverage a pairwise learning approach to rank the candidate keyphrases globally to compare the information saliency between all candidates.
First, we put the candidate keyphrases in the document that are labeled as keyphrases, in the positive set $\mathbf{P}^+$, and the others to the negative set $\mathbf{P}^-$, to obtain the ranking labels.
Then, the loss function is the standard hinge loss in the pairwise learning model:
\begin{equation}
	L_r = \sum_{p^+, p^- \in K} \text{max} (0,  \delta_1 - I_2(p^+) \\+ I_2(p^-)),
\end{equation}
where $I_2(\cdot)$ represents the estimation of information saliency and $\delta_1$ indicates the margin.
It enforces \textit{KIEMP} to rank the candidate keyphrases $p^+$ ahead of $p^-$ within the same document.
Specifically, the information saliency of the $i$-th n-gram representation $c_i^n$ can be computed as follows:
\begin{equation}
	I_2(c_i^n) =\mathbf{W}_2h_i^n + b_2,
\end{equation}
where $\mathbf{W}_2$ is a trainable matrix, and $b_2$ is a bias.
Through the pairwise learning model, we can rank the information saliency of all candidates and extract the keyphrases with more salient information sufficiently.
\subsubsection{Matching for Concept Consistency}
As phrases are used to express various meanings corresponding to different concepts (i.e., topics), a phrase will play different important roles in different concepts of the document \cite{2010topic}.
A matching module is proposed via a metric learning framework to estimate the concept consistency between the candidate keyphrases and their corresponding document.
We first apply variation autoencoder \cite{rezende2014stochastic} on the documents $\mathbf{D}$ and the candidate keyphrases $\mathbf{K}$ to obtain their concepts. 
Each document $D$ is encoded via a latent variable $z \in \mathbb{R}^{c}$ which is assumed to be sampled from a standard Gaussian prior, i.e., $z \sim p({z}) = \mathcal{N}(0,{I}_d)$. Such variable has ability to determine the latent concepts hidden in the documents and will be useful to extract keyphrase \cite{topic}.
During the encoding process, ${z}$ can be sampled via a re-parameterization trick for 
Gaussian distribution, i.e., ${z}\sim q({z}|{D}) = \mathcal{N}({\mu},{\sigma})$. 
Specifically, we sample an auxiliary noise variable ${\varepsilon} \sim N(0,{I})$ and re-parameterization ${z} = {\mu} + {\sigma} \odot \varepsilon$, 
where $\odot$ denotes the element-wise multiplication. The mean vector ${\mu}\in \mathbb{R}^{c}$ and variance vector ${\sigma} \in \mathbb{R}^{c}$ will be inferred by a two-layer network with ReLU-activated function, i.e., 
${\mu} = \mu_{\phi}({D})$ and ${\sigma} = \sigma_{\phi}({D})$ where $\phi$ is the parameter set. 
During the decoding process, the document can be reconstructed by a multi-layer network ($f_{k}$) with Tanh-activated function, i.e., ${\tilde{D}} = f_{k}({z})$. 
Furthermore, the candidate keyphrases are processed in the same way as the documents.

Once having the latent concept representation of the document $z$ and the phrase $z_{i}^n$, the concept consistency can be estimated as follows,
\begin{equation}
	I_3(c_i^n, D) =  z_i^n \mathbf{W}_3z.
\end{equation}
\noindent
Here, $\mathbf{W}_3$ is a learnable mapping matrix. 
The loss function is the triplet loss in the metric learning framework calculated as follows:
\begin{equation}
	L_m=\sum_{p^+, p^- \in K}\text{max} (0, I_3(p^-,D)- I_3(p^+,D) + \delta_2),
\end{equation}
where $\delta_2$ represents the margin.
It enforces \textit{KIEMP} to match and rank the concept consistency of keyphrases $p^+$ ahead of the non-keyphrases $p^-$ within their corresponding document $D$.

Furthermore, to simultaneously minimize the reconstruction loss and penalize the discrepancy between a prior distribution and posterior distribution about the latent variable ${z}$, the VAE process can be implemented by optimizing the following objective function for the documents $L_d$ and the candidate keyphrases $L_k$:
\begin{equation}\label{lossd}
	L_{d} = -\mathbb{E}_{q(\mathbf{z}|\mathbf{D})} \big [p(\mathbf{D}|\mathbf{z}) \big ]+ D_{KL}\big (p(\mathbf{z})||q(\mathbf{z}|\mathbf{D}) \big) ,
\end{equation}
\begin{equation}\label{lossk}
	\setlength\abovedisplayskip{-10pt}
	L_{k} = -\mathbb{E}_{q(\mathbf{z}|\mathbf{K})} \big [p(\mathbf{K}|\mathbf{z}) \big ]+ D_{KL}\big (p(\mathbf{z})||q(\mathbf{z}|\mathbf{K}) \big) ,
\end{equation}
where $D_{KL}$ indicates the Kullback-Leibler divergence between two distributions. And the final loss of this module is calculated as follows:
\begin{equation}\label{losst}
	L_{t} = L_m + \lambda L_{d} + (1-\lambda) L_{k},
\end{equation}
where $\lambda \in (0,1)$ indicates the balance factor.
Through concept consistency matching, we expect to align keyphrases with high-level concepts (i.e., topics or structures) in the document to assist the model in extracting keyphrases with more important concepts.

\subsection{Model Training and Inference}
Multi-task learning has played an essential role in various fields \cite{renzhi}, and has been widely used in the natural language processing tasks \cite{baseline, span2020} recently. Therefore, our framework allows end-to-end learning of syntactic chunking, saliency ranking, and concept matching in this paper.
Then, we define the training objective of the entire framework with the linear combination of $L_c$, $L_r$, and $L_t$:
\begin{equation}
	L = \epsilon_1 L_c + \epsilon_2 L_r + \epsilon_3 L_t,
\end{equation}
where the hyper-parameters $\epsilon_1$, $\epsilon_2$, and $\epsilon_3$ balance the effects of the importance estimation from three perspectives.
Specifically, $\epsilon_1 + \epsilon_2 + \epsilon_3 = 1$.

In this paper, \textit{KIEMP} aims to extract keyphrases according to their saliency. It contains three modules syntactic accuracy chunking, information saliency ranking, and concept consistency matching. Chunking and matching are used to enforce the ranking module to rank the proper candidate keyphrases ahead. Therefore, only the ranking module is used in the inference process (test-phase).

\begin{table}[!thb]
	\small
	\centering
	\renewcommand\tabcolsep{2pt}
	\renewcommand\arraystretch{1.5}
	\begin{tabular}{l|c|c|c}
		\hline \hline
		\multirow{2}{*}{\textit{\textbf{Dataset}}} & \multicolumn{1}{c|}{\small{\textit{Document Len.}}} & \#  \small{\textit{Keyphrase}} & \small{\textit{Keyphrase Len.}} \\ 
		& \small{\textit{Average}}           & \small{\textit{Average}}        &        \small{\textit{Average}}                                 \\ \hline
		\textit{OpenKP}                              & 900.4          & 1.8        & 2.0                                         \\
		\textit{KP20k}                            & 179.8          & 5.3        & 2.0                                          \\ 
		\hline
		\textit{Inspec}                              & 128.7          & 9.8        & 2.5                                         \\
		\textit{Krapivin}                             & 182.6          & 5.8        & 2.2                                       \\ 
		\textit{Nus}                               & 219.1          & 11.7       & 2.2                                         \\ 
		\textit{SemEval}                             & 234.8          & 14.7       & 2.4                                     \\ 
		\hline \hline
	\end{tabular}
	\caption{Statistics of six benchmark datasets. \textit{Document Len.} and \textit{Keyphrase Len.} represent the number of words in the document and keyphrase respectively.}
	\label{dataset}
\end{table}
\section{Experimental Settings}

\subsection{Datasets}
Six benchmark datasets are mainly used in our experiments, \textit{OpenKP} \cite{xiong19}, \textit{KP20k} \cite{catseq17}, \textit{Inspec} \cite{Inspec}, \textit{Krapivin} \cite{Krapivin}, \textit{Nus} \cite{Nus} and \textit{SemEval} \cite{SemEval}. Table~\ref{dataset} summarizes the statistics of each testing sets.

\textbf{\textit{OpenKP}} consists of around 150K documents sampled from the index of the Bing search engine.
In \textit{OpenKP}, we follow the official split of training (134K documents), development (6.6K documents), and testing (6.6K documents) sets.
The keyphrases for each document in \textit{OpenKP} were labeled by expert annotators, with each document assigned 1-3 keyphrases. As a requirement, all the keyphrases appeared in the original document \cite{xiong19}.

\textbf{\textit{KP20k}} contains a large number of high-quality scientific metadata in the computer science domain from various online digital libraries \cite{catseq17}. 
We follow the official setting of this dataset and split the dataset into training (528K documents), validation (20K documents), and testing (20K documents) sets.
From the training set of \textit{KP20k}, we remove all articles that are duplicated in themselves, either in the \textit{KP20k} validation and testing set.
After the cleanup, the \textit{KP20k} dataset contains 504K training samples, 20K validation samples, and 20K testing samples.

To verify the robustness of \textit{KIEMP}, we also test the model trained with \textit{KP20k} dataset on four widely-adopted keyphrase extraction data sets including \textit{Inspec}, \textit{Krapivin}, \textit{Nus}, and \textit{SemEval}.

In this paper, we focus on keyphrase extraction. Therefore, only the keyphrases that appear in the documents are used for training and evaluation. 
 
 \begin{table}[!t]
 	\small
 	\centering
 	\renewcommand\tabcolsep{7pt}
 	\renewcommand\arraystretch{1.5}
 	\begin{tabular}{lc}
 		\hline \hline
 		\textbf{Hyper-parameter} & \textbf{Dimension or Value} \\ 
 		
 		\hline
 		$\lambda$ & $0.5$ \\
 		$\epsilon_1,\epsilon_2,\epsilon_3$ & 1/3 \\
 		$\delta_1, \delta_2$ & 1.0 \\
 		Optimizer & AdamW \\
 		Learning Rate & $1\times10^{-5}$ \\
 		Batch Size & $32$ \\
 		
 		Warm-Up Proportion & $10\%$ \\ 
 		\hline
 		RoBERTa Embedding $(\mathbb{R}^d)$ & 768 \\ 
 		Concept Dimension $(\mathbb{R}^c)$ & 64 \\ 
 		Max Sequence Length & 512 \\ 
 		Maximum Phrase Length $(N)$& 5 \\
 		\hline \hline
 	\end{tabular}
 	\caption{Parameters used for training \textit{KIEMP}.}
 	\label{parameter}
 \end{table}

\begin{table*}[!htb]
	\small
	\centering
	\renewcommand\tabcolsep{7pt}
	\renewcommand\arraystretch{1.5}
	\begin{tabular}{l|ccc|ccc|cc}
		\hline\hline
		\multirow{2}{*}{\normalsize \textbf{\textit{Model}}} & \multicolumn{6}{c|}{\textit{{OpenKP}}}& \multicolumn{2}{c}{\textit{{KP20k}}} \\\cline{2-9} 
		& $R@1$ & $R@3$ & $R@5$  & $F_1@1$ & $F_1@3$ & $F_1@5$   & $F_1@5$ & $F_1@10$  \\ \hline
		\multicolumn{9}{l}{\textit{{Unsupervised Methods}}}\\\hline
		\multicolumn{1}{l|}{\textit{TFIDF \cite{tfidf}}} 
		& \underline{0.150} & \underline{0.284} & \underline{0.347} & \underline{0.196}* & \underline{0.223}* & \underline{0.196}* & 0.105 & 0.130 \\
		
		\multicolumn{1}{l|}{\textit{TextRank \cite{textrank}}} 
		& 0.041 & 0.098 & 0.142 & 0.054* & 0.076* & 0.079* & \underline{0.180} & \underline{0.150} \\
		
		\hline
		\multicolumn{9}{l}{\textit{{Supervised Methods with Additional Features}}}\\\hline
		
		\multicolumn{1}{l|}{\textit{BLING-KPE \cite{xiong19}}} 
		& 0.220 & 0.390 & 0.481 & 0.285* & 0.303* & 0.270* & - & - \\

		\multicolumn{1}{l|}{{\textit{SMART-KPE+R2J \cite{2020sota}}}} 
		& \underline{0.307} & \underline{0.532} & \underline{0.625} & \underline{0.381} & \underline{0.405} & \underline{0.347} & -  & -  \\
		\hline
		
		\multicolumn{9}{l}{\textit{{Supervised Methods without Additional Features}}}\\\hline
		\multicolumn{1}{l|}{\textit{CopyRNN \cite{catseq17}}} 
		& 0.174 & 0.331 & 0.413 & 0.217* & 0.237* & 0.210* & 0.327 & 0.278  \\
		
		\multicolumn{1}{l|}{\textit{DivGraphPointer \cite{gcn2019}}} 
		& - & - & - & - & - & - & 0.368 & 0.292  \\
		
		\multicolumn{1}{l|}{\textit{Div-DGCN \cite{gcn2020}}} 
		& - & - & - & - & - & - & 0.349 & 0.313  \\
		
		\multicolumn{1}{l|}{\textit{SKE-Large-CLS \cite{span2020}}} 
		& - & - & - & - & - & - & 0.392 & 0.330  \\
		
		\multicolumn{1}{l|}{{\textit{ChunkKPE \cite{baseline}}}} 
		& {0.283} & {0.486} & 0.581 & 0.355 & 0.373 & 0.324 & 0.408  & 0.337  \\
		
		\multicolumn{1}{l|}{{\textit{RankKPE \cite{baseline}}}} 
		& {0.290} & {0.509} & 0.604 & 0.361 & 0.390 & 0.337 & 0.417  & 0.343  \\
		
		\multicolumn{1}{l|}{{\textit{JointKPE \cite{baseline}}}} 
		& {0.291} & {0.511} & 0.605 & 0.364 & 0.391 & 0.338 & 0.419  & 0.344  \\
		
		\multicolumn{1}{l|}{{{\textit{\textbf{KIEMP}}}}}
		& \textbf{0.298} & \textbf{0.517} & \textbf{0.615} & \textbf{0.369} & \textbf{0.392} & \textbf{0.340} & \textbf{0.421}  & \textbf{0.345}  \\ 
		
		\hline\hline
	\end{tabular}
	\caption{Performances of keyphrase extraction model on the \textit{OpenKP} development set and the \textit{KP20k} testing set. The best results of our model are highlighted in bold, and the best results of baselines are underlined. * indicates these numbers are not included in the original paper and are estimated with Precision and Recall. The results of the baselines are reported in their corresponding papers.}
	\label{overall}
\end{table*}
\subsection{Baselines}
This paper focuses on the comparisons with the state-of-the-art baselines and chooses the following keyphrase extraction models as our baselines.

\mytext{TextRank} An unsupervised algorithm based on weighted-graphs proposed by \citet{textrank}. Given a word graph built on co-occurrences, it calculates the importance of candidate words with PageRank. The importance of a candidate keyphrase is then estimated as the sum of the scores of the constituent words.

\mytext{TFIDF} \cite{tfidf} is computed based on candidate frequency in the given text and inverse document frequency

\mytext{CopyRNN} \cite{catseq17} which uses the attention mechanism as the copy mechanism to extract keyphrases from the given document.

\mytext{BLING-KPE} \cite{xiong19} first concatenates the pre-trained language model (ELMo \cite{elmo}) as word embeddings, visual as well as positional features, and then uses a CNN network to obtain n-gram phrase embeddings for binary classification.

\mytext{JointKPE} \cite{baseline} jointly learns a chunking model (\textit{ChunkKPE}) and a ranking model (\textit{RankKPE}) for keyphrase extraction.

\mytext{SMART-KPE+R2J} \cite{2020sota} presents a multi-modal method to the keyphrase extraction task, which leverages lexical and visual features to enable strategy induction as well as meta-level features to aid in strategy selection.

\mytext{DivGraphPointer} \cite{gcn2019} combines the advantages of traditional graph-based ranking methods and recent neural network-based approaches. Furthermore, they also propose a diversified point network to generate a set of diverse keyphrases out of the word graph in the decoding process.

\mytext{Div-DGCN} \cite{gcn2020} proposes to adopt the Dynamic Graph Convolutional Networks (DGCN) to acquire informative latent document representation and better model the compositionality of the target keyphrases set.

\mytext{SKE-Large-CLS} \cite{span2020} obtains span-based representation for each keyphrase and further learns to capture the similarity between keyphrases in the source document to get better keyphrase predictions.

In this paper, for ease of introduction, all the baselines are divided according to the following three perspectives, syntax, saliency, and combining syntax and saliency. Among them,  \textit{BLING-KPE}, \textit{CopyRNN},\textit{ ChunkKPE} belong to the former, \textit{TFIDF}, \textit{TextRank}, as well as \textit{RankKPE} belong to the second, and \textit{DivGraphPointer}, \textit{Div-DGCN}, \textit{SKE-Large-CLS}, \textit{SMART-KPE+R2J}, and \textit{JointKPE} belong to the last.
\subsection{Evaluation Metrics}
For the keyphrase extraction task, the performance of keyphrase model is typically evaluated by comparing the top $k$ predicted keyphrases with the target keyphrases (ground-truth labels).
The evaluation cutoff $k$ can be a fixed number (e.g., $F_1@5$ compares the top-$5$ keyphrases predicted by the model with the ground-truth to compute an $F_1$ score).
Following the previous work \cite{catseq17, gcn2019}, we adopt macro-averaged recall and F-measure ($F_1$) as evaluation metrics, and $k$ is set to be 1, 3, 5, and 10.
In the evaluation, we apply Porter Stemmer \cite{stemmer} to both target keyphrases and extracted keyphrases when determining the match of keyphrases and match of the identical word.
\subsection{Implementation Details}

Implementation details of our proposed models are as follows.
The maximum document length is 512 due to BERT limitations \cite{bert}, and documents are zero-padded or truncated to this length.
The training used 4 GeForce RTX 2080 Ti GPUs and took about 31 hours and 77 hours for \textit{OpenKP} and \textit{KP20k} datasets respectively.
Table~\ref{parameter} lists the parameters of our model.
Furthermore, the model was implemented in Pytorch \cite{pytorch} using the huggingface re-implementation of RoBERTa \cite{transformer_pytorch}.

\section{Results and Analysis}

This section investigates the performance of the proposed \textit{KIEMP} on six widely-used benchmark datasets (\textit{OpenKP}, \textit{KP20k}, \textit{Inspec}, \textit{Krapivin}, \textit{Nus}, and \textit{Semeval}) from three facets. The first one demonstrates its superiority by comparing it with ten baselines in terms of several metrics. The second one is to verify the sensitivity of the concept dimension. The last one is to explicitly show the keyphrase extraction results of \textit{KIEMP} via two examples (two testing documents).

\subsection{Overall Performance}
The overall performance of different algorithms on two benchmarks (\textit{OpenKP} and \textit{KP20k}) is summarized in Table~\ref{overall}.
We can see that the supervised methods outperform all the unsupervised algorithms (\textit{TFIDF} and \textit{TextRank}). This is not surprising since the supervised methods are trained end-to-end with supervised data.
In all the supervised baselines, the methods using additional features are better than those without additional features. 
The reason is that the models with additional features are equal to encode keyphrases from multiple features perspectives. Therefore, it is helpful for the model to measure the importance of each keyphrase, thus improving the performance of the result of keyphrase extraction.
Intuitively, this is the same as our proposed method.
\textit{KIEMP} considers the importance of keyphrases from multiple perspectives and fairly measures the importance of each keyphrase. 
But the difference is that we do not need additional features to assist. 
And in many practical applications of keyphrase extraction, there is no additional feature (i.e., visual features) information to use in most cases.
Compared with recent baselines (\textit{ChunkKPE}, \textit{RankKPE}, and \textit{JointKPE}), \textit{KIEMP} performs stably better on all metrics on both two datasets. These results demonstrate the benefits of estimating the importance of keyphrases from multiple perspectives simultaneously and the effectiveness of our multi-task learning strategy.

\begin{figure}
	\centering
	\includegraphics[scale=0.49]{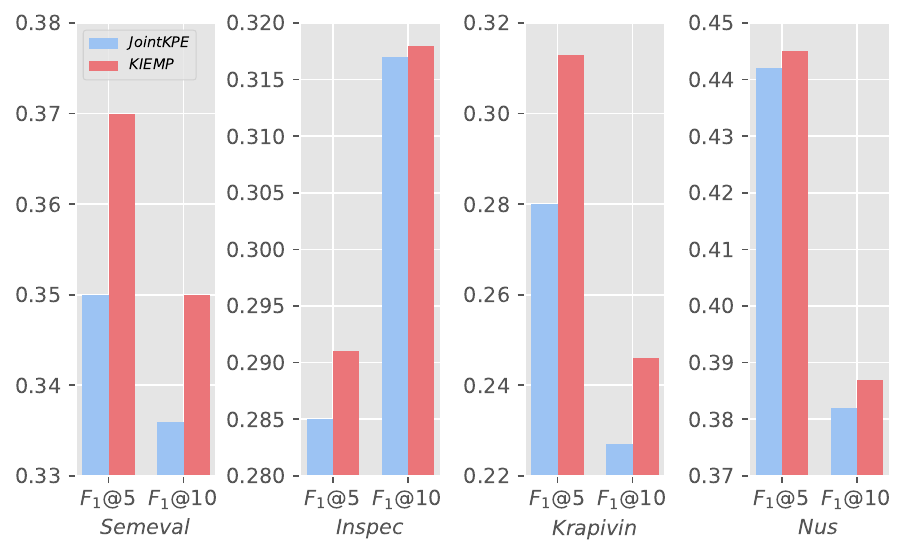}
	\caption{Results of keyphrase extraction model on four testing sets (\textit{Semeval}, \textit{Inspec}, \textit{Krapivin}, and \textit{Nus}). The results of \textit{JointKPE} are re-evaluated via the code which is provided by its corresponding paper.}
	\label{test}
\end{figure}

Furthermore, to verify the robustness of \textit{KIEMP}, we also test the \textit{KIEMP} trained with \textit{KP20k} dataset on four widely-adopted keyphrase extraction data sets.
It can be seen from Figure~\ref{test} that \textit{KIEMP} is superior to the best baseline (\textit{JointKPE}). We consider that this phenomenon comes from two benefits.
One is that the high-level concepts captured by a deep latent variable model may contain topic and structure features. These features are essential information to evaluate the importance of phrases.
Another one is that the latent variable is characterized by a probability distribution over possible values rather than a fixed value, which can enforce the uncertainty of our model and further lead to robust representation learning.

\begin{table}[!htb]
	\small
	\centering
	\renewcommand\tabcolsep{5pt}
	\renewcommand\arraystretch{1.5}
	\begin{tabular}{l|ccc}
		\hline\hline
		\multirow{2}{*}{ \textbf{\textit{Concept Dimension $(\mathbb{R}^c)$}}} & \multicolumn{3}{c}{\textit{{OpenKP}}} \\ 
		& $R@1$ & $R@3$ & $R@5$ \\ \hline
		\multicolumn{1}{c|}{\textit{64 }}
		& \textbf{0.298} & \textbf{0.517} & \textbf{0.615}\\
		\multicolumn{1}{c|}{\textit{256 }}
		& 0.297 & 0.513 & 0.610  \\ 
		\multicolumn{1}{c|}{\textit{512 }}
		& 0.296 & 0.509 & 0.609  \\
		\multicolumn{1}{c|}{\textit{768 }}
		& 0.293 & 0.508 & 0.606  \\
		\hline\hline
	\end{tabular}
	\caption{Effectiveness of different dimensions of latent concept representation. The best results are highlighted in bold.}
	\label{concept}
\end{table}

\begin{table*}[!htb]
	\renewcommand\arraystretch{1.7}
	\begin{center}
		\small
		\begin{tabular}{|p{15cm}|}
			\hline
			(A) \textbf{\textit{Part of the Input Document}}: \\
			
			\small
			The Great Plateau is a large region of land that is secluded from other parts of Hyrule, as its steep slopes prevent anyone from traveling to and from it without special equipment, such as the Paraglider. The only active inhabitant is the Old Man, a mysterious ... 
			(URL: https://zelda.gamepedia.com/Great\_Plateau)
			\\ 
			\textbf{\textit{Target Keyphrase}}: 
			
			(1) \mytext{\textcolor{magenta}{great plateau}} ; (2) \mytext{\textcolor{magenta}{breath of the wild}} ; (3) \mytext{\textcolor{magenta}{hyrule}}\\

			
			
			\hline
			
			\textbf{\textit{KIEMP without concept consistency matching}}: 
			
			(1) \mytext{\textcolor{magenta}{great plateau}} ; (2) \mytext{\textcolor{magenta}{hyrule}} ; (3) \mytext{\textcolor{magenta}{breath of the wild}} ; (4) paraglider ; (5) zelda 
			\\

			

			\textbf{\textit{KIEMP}}: 
			
			(1) \mytext{\textcolor{magenta}{great plateau}} ; (2) \mytext{\textcolor{magenta}{breath of the wild}} ; (3) \mytext{\textcolor{magenta}{hyrule}} ; (4) paraglider ; (5) starting region 
			\\\hline\hline

			(B) \textbf{\textit{Part of the Input Document}}: \\
			
			\small
			Transformational leaders also depend on visionary leadership to win over followers, but they have an added focus on employee development. For example, a transformational leader might explain how her plan for the future serves her employees’ interests and how she will support them through the changes ... 
			
			(URL: https://yourbusiness.azcentral.com/managers-different-leadership-styles-motivate-teams-8481.html)
			\\ 
			\textbf{\textit{Target Keyphrase}}: 
			
			(1) \mytext{\textcolor{magenta}{managers}} ; (2) \mytext{\textcolor{magenta}{leadership}} ; (3) \mytext{\textcolor{magenta}{teams}}
			\\
			\hline
			
			
			
			\textbf{\textit{KIEMP without concept consistency matching}}: 
			
			(1) motivating ; (2) motivate ; (3) charismatic leadership ; (4) transformational leadership ; (5) employee development
			\\

			

			\textbf{\textit{KIEMP}}: 
			
			(1) leadership styles; (2) \mytext{\textcolor{magenta}{managers}} ; (3) charismatic leadership ; (4) transformational leadership ; (5) \mytext{\textcolor{magenta}{leadership}} 
			\\\hline
			
		\end{tabular}
		\caption{\label{case} Example of keyphrase extraction results (selected from the \textit{OpenKP} dataset). Phrases in red and bold are target keyphrases predicted by the different models (\textit{KIEMP} without concept consistency matching and \textit{KIEMP}).}
	\end{center}
\end{table*}

\subsection{Sensitivity of the Concept Dimension}
Here, we verify the effectiveness of using different concept dimensions.
From Table~\ref{concept}, we can find that the increase of the dimension of latent concept representation has little effect on the result of keyphrase extraction.
In contrast, the smaller the dimension, the better the result. 
Furthermore, in Table~\ref{overall}, the improvement of our proposed \textit{KIEMP} model on the $F_1@1$ evaluation metric is higher than the $F_1@3$ and $F_1@5$ evaluation metrics on the \textit{OpenKP} dataset.
We consider the main reason is that our concept representation may capture the high-level conceptual information of phrases or documents, such as topics and structure information. 
Therefore, \textit{KIEMP} with concept consistency matching module focuses more on extracting keyphrases closest to the main topic of the given document.

\subsection{Case Study}
To further illustrate the effectiveness of the proposed model, we present a case study on the results of the keyphrases extracted by different algorithms.
Table~\ref{case} presents the results of \textit{KIEMP} without concept consistency matching and \textit{KIEMP}.
From the first example, we can see that our \textit{KIEMP} model is more inclined to extract keyphrases closer to the central semantics of the input document, which benefits from our concept consistency matching model.
From the second example, we can see that the keyphrases extracted by \textit{KIEMP} without concept consistency matching contain some redundant or meaningless phrases.
The main reason may be that the \textit{KIEMP} without concept consistency matching does not measure the importance of phrases from multiple perspectives, which leads to biased extraction.
On the contrary, the keyphrases extracted by \textit{KIEMP} are all around the main concepts of the example document, i.e., ``leadership''. It further demonstrates the effectiveness of our proposed model.

\section{Conclusions and Future Work}

A new keyphrase importance estimation from the multiple perspectives approach is proposed to estimate the importance of keyphrase. Benefiting from the designed syntactic accuracy chunking, information saliency ranking, and concept consistency matching modules, \textit{KIEMP} can fairly extract keyphrases. A series of experiments have demonstrated that \textit{KIEMP} outperformed the existing state-of-the-art keyphrase extraction methods.
In the future, it will be interesting to introduce an adaptive approach in \textit{KIEMP} to filter the meaningless phrases.

\section{Acknowledgments}
This work was supported in part by the National Key Research and Development Program of China under Grant 2020AAA0106800; the National Science Foundation of China under Grant 61822601 and 61773050; the Beijing Natural Science Foundation under Grant Z180006; The Fundamental Research Funds for the Central Universities (2019JBZ110).

\bibliography{anthology,custom}
\bibliographystyle{acl_natbib}
\end{document}